\documentclass{article}

\usepackage{amsmath}
\usepackage{authblk}
\usepackage{graphicx}
\usepackage{multirow}
\usepackage{booktabs}
\usepackage{float}
\usepackage{tabularx}
\usepackage{colortbl}
\usepackage{rotating}
\usepackage{threeparttable}

\usepackage[table]{xcolor} 
\usepackage[ruled,vlined,noend,nofillcomment]{algorithm2e}
\usepackage[colorlinks=true, allcolors=blue]{hyperref}
\usepackage[letterpaper,top=2cm,bottom=2cm,left=3cm,right=3cm,marginparwidth=1.75cm]{geometry}

\usepackage{lineno}
\usepackage[backend=biber,style=nature,citestyle=numeric-comp,sorting=none]{biblatex}
\DeclareFieldFormat[inproceedings]{title}{#1}  
\hypersetup{
    colorlinks=true, 
    linkcolor=red,  
    citecolor=blue, 
    filecolor=magenta, 
    urlcolor=cyan   
}

\definecolor{highlightcolor}{rgb}{0.88,1,1} 
\definecolor{gray}{rgb}{0.9,0.9,0.9}
\definecolor{orange}{rgb}{1.0,0.92,0.8}

\title{Rejoining fragmented ancient bamboo slips with physics-driven deep learning}

\author[1]{Jinchi Zhu$^\dagger$}
\author[5]{Zhou Zhao$^\ddagger$}
\author[2]{Hailong Lei$^\S$}
\author[3]{Xiaoguang Wang}
\author[2]{Jialiang Lu}
\author[2]{Jing Li}
\author[1]{Qianqian Tang}
\author[1]{Jiachen Shen}
\author[4]{Gui-Song Xia$^*$}
\author[1,4]{Bo Du$^*$}
\author[1,4]{Yongchao Xu$^*$}

\affil[1]{School of Computer Science, Wuhan University}
\affil[2]{Center of Bamboo and Silk Manuscripts, Wuhan University}
\affil[3]{School of Information Management, Wuhan University}
\affil[4]{School of Artificial Intelligence, Wuhan University}
\affil[5]{School of Computer Science, Central China Normal University}

\begin{document}
\date{}
\maketitle

\begin{abstract}
Bamboo slips are a crucial medium for recording ancient civilizations in East Asia, and offers invaluable archaeological insights for reconstructing the Silk Road, studying material culture exchanges, and global history. However, many excavated bamboo slips have been fragmented into thousands of irregular pieces, making their rejoining a vital yet challenging step for understanding their content. Here we introduce WisePanda, a physics-driven deep learning framework designed to rejoin fragmented bamboo slips. Based on the physics of fracture and material deterioration, WisePanda automatically generates synthetic training data that captures the physical properties of bamboo fragmentations. This approach enables the training of a matching network without requiring manually paired samples, providing ranked suggestions to facilitate the rejoining process. Compared to the leading curve matching method, WisePanda increases Top-50 matching accuracy from 36\% to 52\% among more than one thousand candidate fragments. Archaeologists using WisePanda have experienced substantial efficiency improvements (approximately 20 times faster) when rejoining fragmented bamboo slips. This research demonstrates that incorporating physical principles into deep learning models can significantly enhance their performance, transforming how archaeologists restore and study fragmented artifacts. WisePanda provides a new paradigm for addressing data scarcity in ancient artifact restoration through physics-driven machine learning.
\end{abstract}

\begin{refsection}
\section*{Main}
\begin{figure}[h!]
\centering
\includegraphics[width=1\linewidth]{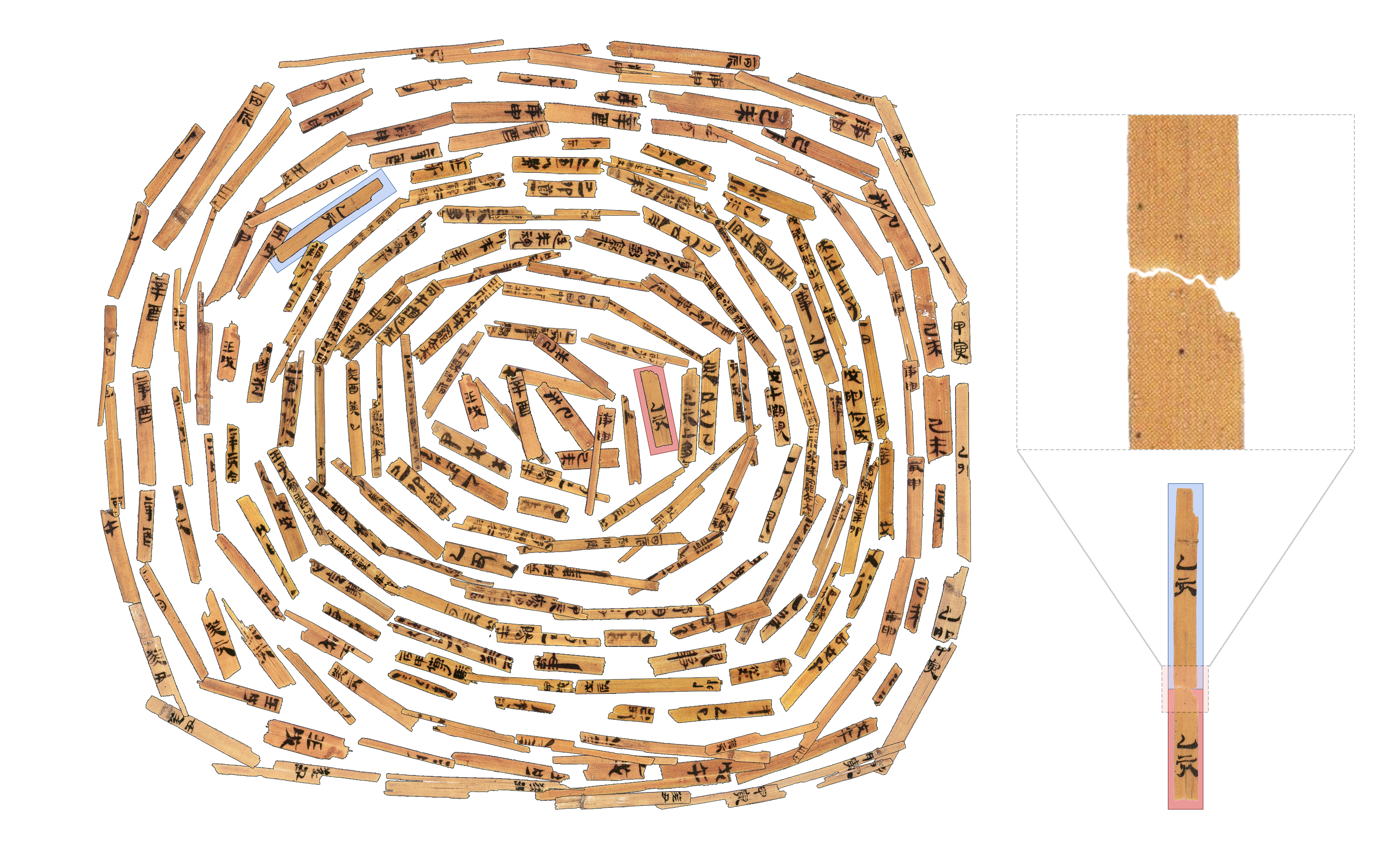}
\caption{\textbf{The challenge of bamboo slip rejoining in archaeological research.} Some excavated bamboo slip fragments from the Qin dynasty (221--206 BCE) arranged in a spiral pattern (left) illustrate the overwhelming scale of the rejoining challenge faced by archaeologists---with thousands of fragments at single excavation sites, each potentially matching with any other piece. A zoomed view of a representative fragment (upper right) reveals the characteristic irregular breakage pattern along the fracture curve. An example of a reconstructed complete slip (lower right) demonstrates how proper rejoining establishes morphological continuity, which is essential for the interpretation of ancient texts. The complex morphology of the fractures, coupled with deterioration from millennia underground, creates a matching problem of extraordinary complexity that has traditionally required intensive manual effort, with experts sometimes spending weeks to successfully match a single pair of fragments.}
\label{fig:bamboo_rejoining_challenge}
\end{figure}

Bamboo slips, serving as a fundamental medium for documenting ancient East Asian civilizations, contain invaluable historical records spanning philosophy, law, and social life of the period~\cite{allan2015buried, liu2024approaching, xiaogan2003bamboo, ling1990formulaic, zheng2006brief}. Their durability has enabled these artifacts to survive millennia underground while retaining legible content, offering scholars unprecedented insights into historical societies. However, the excavation of these delicate artifacts presents a critical challenge~\cite{johnson2019archaeological, liu2024approaching} - many bamboo slips have been fragmented into thousands of pieces (Figure~\ref{fig:bamboo_rejoining_challenge}), significantly complicating efforts to reconstruct and interpret their content. This fragmentation creates a fundamental obstacle in accessing the wealth of historical knowledge contained in these artifacts.

The rejoining of fragmented bamboo slips represents one of the most challenging problems in archaeological preservation and cultural heritage studies~\cite{zhou2017identifying}. The difficulty stems from multiple factors: First, the enormous number of potential fragment combinations makes manual matching extremely time-consuming - for instance, The Qin bamboo slips from the Shuihudi site comprise fragments numbering in the tens of thousands, with each piece potentially matching any of the others~\cite{Shuihudi1990QinSlips}. Second, environmental factors like moisture and pressure have caused extensive physical deterioration~\cite{cha2014micromorphological}, distorting the original shapes and surfaces. 
Third, the sparse recorded characters often make it difficult to capture textual remnants at the fracture curves created by transverse breakage, rendering text-based matching approaches largely ineffective and necessitating reliance primarily on the morphological patterns of fracture curves for identification. The unique fiber structure of bamboo creates complex fracture patterns that traditional curve-matching approaches struggle to analyze effectively. These challenges have made the restoration process highly labor-intensive, with experts sometimes requiring weeks to successfully match a single pair of fragments.

\subsection*{Deep learning for ancient artifact repair}
Here we overcome the challenges of fragmented bamboo slip rejoining through physics-driven deep learning. Our approach integrates principles from the physics of fracture with deep neural networks, enabling automated matching of fragmented pieces without relying on manually labeled training data. Deep learning~\cite{lecun2015deep} has been successfully used in restoring and attributing ancient texts~\cite{assael2022restoring} empowered by large training datasets. Beyond cultural heritage applications, deep learning has also demonstrated remarkable success across diverse scientific domains~\cite{silver2017mastering, senior2020improved, jumper2021highly, dafoe2021cooperative, merchant2023scaling, lu2024multimodal, bodnar2025foundation}. While recent years have also seen significant progress in fragmented cultural relic restoration~\cite{bickler2021machine} with labeled data - from oracle bone reconstruction~\cite{meng2018recognition, zhang2022data} to manuscript fragment matching~\cite{zhu2007globally, zheng2024reunion} - most existing approaches rely on traditional computer vision techniques and struggle with complex degradation patterns~\cite{li2014investigation}. Some works have attempted to use generative models for cultural heritage restoration~\cite{zhang2022data}. While achieving promising results, these methods are constrained by the scarcity of training data and often fail to capture the intricate physical properties of ancient materials~\cite{eslami2020review,hou2018novel}.

\begin{figure}
\centering
\includegraphics[width=1\linewidth]{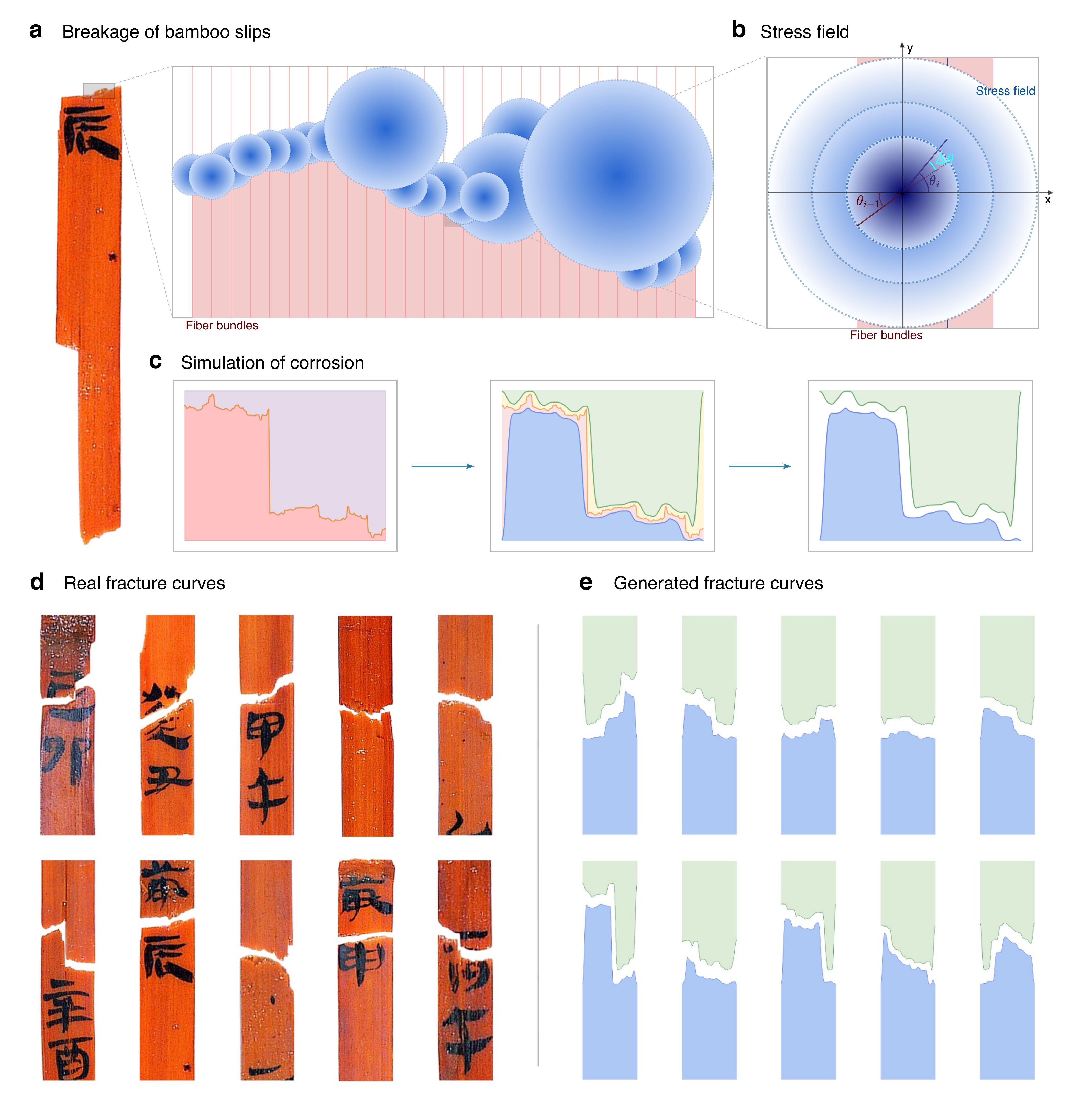}
\caption{\textbf{Physics-driven modeling of bamboo slip fracture and deterioration.} \textbf{a}, The breakage process of bamboo slips showing how fracture propagates across the bamboo's fiber structure, with the resulting irregular curve composed of black line segments and the corresponding stress field distribution (blue gradient). \textbf{b}, Detailed stress field model illustrating the mathematical relationship between the fracture angles ($\theta_{i-1}$, $\theta_i$) and the stress propagation in the $x$-$y$ coordinate system, where the blue dotted circles represent the stress field emanating from the fracture endpoint, showing how stress radiates outward and concentrates at fiber boundaries. \textbf{c}, Time-sequential simulation of the corrosion process, demonstrating how the original fracture pattern (left) changes through environmental exposure (middle) to produce the final deteriorated curve morphologies (right), with protruding areas experiencing accelerated degradation. \textbf{d}, Collection of real bamboo slip fracture curves extracted from archaeological samples, exhibiting diverse breakage patterns that serve as reference for model validation. \textbf{e}, Synthetically generated fracture curves produced by our physics-driven model, displaying morphological characteristics highly similar to the real samples in panel d. This approach enables the generation of extensive paired training data that captures both the physical properties of bamboo fragmentation and the effects of long-term degradation, effectively addressing the data scarcity challenge inherent in ancient artifact restoration.}
\label{fig:physics_model}
\end{figure}

We propose WisePanda, a deep learning framework that leverages physical principles to overcome the data scarcity problem inherent in fragmented bamboo slip rejoining task. Training WisePanda presents a unique paradox: while manual fragment rejoining is prohibitively time-consuming. The very problem we aim to solve - this same process would traditionally be required to generate training data for the model. We resolve this dilemma by resorting to the physics of fracture~\cite{anderson2005fracture}. By modeling the physical properties of bamboo and the process that govern its degradation~\cite{tan2011mechanical,cha2014micromorphological}, we generate extensive synthetic training data that captures the essential characteristics of real paired fragment slips (Figure~\ref{fig:physics_model}). This physics-driven approach enables us to produce large-scale, realistic paired training data without requiring manual matching efforts, while ensuring the model learns meaningful patterns based on actual material properties rather than superficial features.

\subsubsection*{Data synthesis with physics of fracture}
The generation of training data for bamboo slip rejoining requires a comprehensive understanding of both fracture formation and degradation processes~\cite{behera2025review}. Our physical model builds on two key observations: First, bamboo slips exhibit distinct ``transverse'' and ``longitudinal'' fracture modes due to their unique fiber structure~\cite{behera2025review}, as illustrated in Extended Figure~\ref{fig:bamboo_slips_fractures}, with transverse fractures accounting for approximately 70\% of cases. Second, these fractures undergo complex corrosion processes over time, significantly altering their original patterns~\cite{chen2022state, cha2014micromorphological, liu2024approaching}. Building on these insights, we develop a systematic approach to generate synthetic fragment pairs for transverse fractures, which represent the majority of rejoining challenges.

The physics of fracture in bamboo slips is governed by their distinctive vertical fiber arrangement~\cite{dixon2014structure, ramful2020investigation, chen2019fracture}, as detailed in Extended Figure~\ref{fig:physical_model}. When stress extends horizontally across the slip, it creates characteristic fracture patterns as it propagates through consecutive fiber bundles~\cite{chen2019fracture, ramful2022investigation}. We model this process through a probabilistic framework where the fracture angle of each fiber bundle is influenced by the stress field generated at the fracture endpoint of the preceding fiber bundle (Figure~\ref{fig:physics_model}a and b). This physical relationship can be expressed through a probability density function that determines likely fracture paths based on stress distribution and geometric configurations. The corrosion process is then simulated by analyzing geometric exposure - protruding areas and exposed surfaces are more susceptible to degradation through moisture absorption and microorganism activity~\cite{blanchette1994assessment}. To efficiently calculate the exposure area of each fiber bundle, we ingeniously employ ReLU functions that compare the height differences between adjacent fiber bundles, effectively capturing only the protruding portions that are more vulnerable to environmental deterioration (Figure~\ref{fig:physics_model}c). This computational approach allows us to precisely model the progressive and differential deterioration patterns observed in archaeological samples.

To ensure the generated data accurately reflects real bamboo slip characteristics, we optimize model parameters using a genetic algorithm~\cite{Goldberg1989}. The algorithm evaluates parameter sets by comparing the distribution of generated fragments against a reference set of 200 real fragment curves through dimensionality reduction and Silhouette analysis~\cite{van2008visualizing}. Our analysis shows high similarity between the synthetic and real data distributions. The resulting pipeline enables us to generate extensive paired training data (Figure~\ref{fig:physics_model}d and e) that captures both the physical properties of fracture formation and the effects of long-term degradation, effectively addressing the data scarcity challenge in rejoining fragmented bamboo slips.

\subsubsection*{WisePanda is a ranking system}
\begin{figure}[h!]
\centering
\includegraphics[width=1\linewidth]{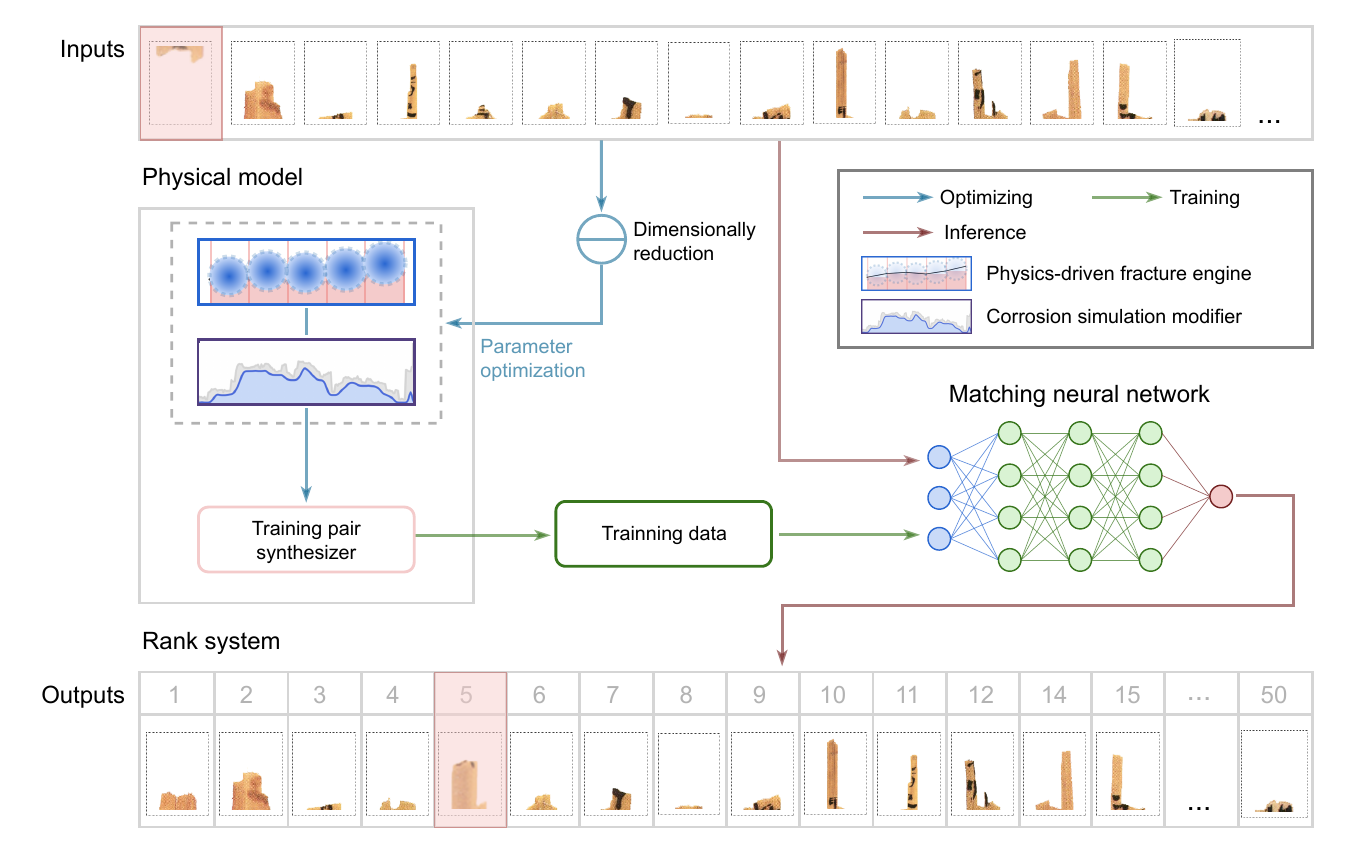}
\caption{\textbf{Framework of WisePanda integrating physics-driven deep learning for bamboo slip rejoining.} The workflow diagram illustrates the complete pipeline of our system. The process begins with input bamboo slip fragments (top row), where the target fragment is highlighted in pink. In the physical model stage, we implement two key components: a physics-driven fracture engine that generates initial breakage patterns, and a corrosion simulation modifier that replicates long-term degradation effects. The dimensionality reduction module (center) enables comparison between real and synthetic fragments through statistical visualization techniques, facilitating parameter optimization to ensure realistic fracture simulation. The optimized physical model feeds into the training pair synthesizer, which generates matching fragment pairs for training the matching network. This matching network learns to effectively represent and compare fragment curves through a multi-layer architecture with specialized embedding capabilities. During inference (red arrows), the system compares the target fragment against all candidates, producing a ranked list of potential matches (bottom row). The ranking transforms thousands of possible matches into a manageable set, with the system highlighting the most probable candidates (position 5 shown in pink represents a correct match). This approach enables archaeologists to focus their verification efforts on the most promising candidates, significantly accelerating the traditionally time-consuming rejoining process.}
\label{fig:wisepanda_framework}
\end{figure}

WisePanda's architecture features an interpretable pipeline for fragmented bamboo slip rejoining task (Figure~\ref{fig:wisepanda_framework}), using synthesized data with physics of fracture for model training and providing Top-k candidates to assist archaeologists. At its core lies a TripletNet-based deep learning network~\cite{hoffer2015deep}, designed to learn effective feature representation that can distinguish between matching and non-matching fragments. The network processes each input as a 64-dimensional vector that captures the detailed features of the fracture curves. By leveraging a triplet loss function~\cite{hoffer2015deep, tian2020hynet}, the network learns to minimize the distance between matching pairs while maximizing the distance between non-matching ones.

In practical deployment, WisePanda operates by predicting matching probabilities between fragments and offers archaeologists a ranked list of the top potential matches for each fragment. This significantly reduces the search space from thousands of candidates to a manageable set~\cite{funkhouser2011learning, brown2012tools}, thereby streamlining the rejoining process. The network's output is a value between 0 and 1, indicating the match probability between a given pair of fragments, where a value closer to 1 signifies a higher likelihood of a correct match. This ranking approach effectively mitigates the overwhelming task of manually matching fragments by narrowing down the possibilities to the most probable matches.

\subsubsection*{WisePanda as an archaeologist adjunct}
To bridge the gap between computational methods and archaeological research practices, we have developed an intuitive computer-assisted tool that implements WisePanda's capabilities in a practical workflow (Extended Figure~\ref{fig:software}). When working with this tool, archaeologists first select a query fragment from the collection. The system then automatically generates a ranked list of potential matching candidates, significantly narrowing down the search space from thousands to dozens of fragments. Each suggested match is accompanied by a matching score given by WisePanda's matching network, allowing archaeologists to prioritize their examination of the most promising candidates.

The tool provides a dedicated workspace where archaeologists can visually manipulate and compare fragments through intuitive operations such as dragging, rotating, and fine-tuning alignments. To facilitate detailed examination, the system offers specialized visualization features like layer swapping and curve enhancement, enabling careful inspection of  fracture patterns, ink traces, and textures (\textit{e.g.}, fiber patterns). All successful matches are saved and documented in a centralized database, creating a growing repository of verified rejoining that contributes to the broader restoration effort. This computer-assisted tool has significantly accelerated the traditional rejoining workflow~\cite{brown2012tools}, ensuring archaeologists maintain full control over the final verification of matches while benefiting from AI-powered suggestions.

\subsection*{Experimental evaluation}
\noindent\textbf{Datasets.} For evaluating WisePanda's effectiveness, we collected several datasets of bamboo and wooden slip fragments. Our primary test set (Bamboo236) consists of 118 pairs of matched bamboo slip fragments, carefully verified and provided by archaeological experts who successfully rejoined these fragments through meticulous manual work (achieved by three archaeological experts within half a year). These bamboo slip artifacts are part of an ongoing archaeological research project, with the cultural relics yet to be formally published. To simulate more challenging real-world scenarios, we created an extended dataset (Bamboo1350) by introducing 1114 additional interference fragments from the same set, expanding the candidate pool to 1,350 total fragments while maintaining the original matched pairs. To assess generalization capabilities across different materials, we also collected wooden slip datasets. The base wooden slip test set (Wood670) contains 335 expert-verified pairs, while its extended version (Wood3833) includes 3163 additional interference fragments. These diverse datasets enable comprehensive evaluation of fragment matching methods under varying conditions and material types.

\medskip
\noindent\textbf{Evaluation methodology.} The evaluation process involved computing similarity scores between each fragment's fracture curve and all potential matching candidates from the pool. Fragments were pre-classified as upper or lower parts, with matching only performed between these distinct groups. We extracted features from both fracture curves of each fragment, ranked the computed similarity scores, and measured the probability of finding the correct match within the Top-k predictions. We conducted comprehensive experiments comparing our WisePanda with several classical methods, including Manual Random Search, curve matching techniques (using Dynamic Time Warping (DTW))~\cite{berndt1994using}, Fast Matching Method (FMM)~\cite{frenkel2003curve}, and Scale Invariant Signature (SIS)~\cite{calabi1998differential}. This comparative framework allows us to assess the relative performance of our physics-driven approach against established techniques in the field of fragment matching.

\medskip
\noindent\textbf{Results and analysis.} The performance variations among different methods reveal insights about the nature of the fragmented bamboo slip matching problem. In our original test set (Bamboo236), WisePanda demonstrated superior performance across all metrics, achieving a Top-1 accuracy of 12.29\%, Top-5 accuracy of 35.17\%, Top-10 accuracy of 52.54\%, Top-20 accuracy of 69.07\%, and Top-50 accuracy of 94.07\%. Among traditional methods, SIS~\cite{calabi1998differential} achieved the second-best performance with a Top-1 accuracy of 10.59\% and Top-50 accuracy of 72.88\%, followed by DTW~\cite{berndt1994using} with a Top-50 accuracy of 75.42\%. FMM~\cite{frenkel2003curve} performed less effectively, with Top-50 accuracies of 65.25\%. To simulate real-world archaeological scenarios where numerous unrelated fragments exist, we expanded our test dataset by introducing additional interference fragments. As shown in Extended Figure~\ref{fig:extended_robustness}, when the candidate pool was increased by a factor of approximately 10, WisePanda maintained a Top-50 accuracy of 52.54\% compared to the original 94.07\%, demonstrating robust performance even with significantly increased search space.

\begin{table}[htbp]
\centering
\small
\begin{threeparttable}
\caption{Comparison of fragment matching methods across different test scenarios}
\label{tab:comparative_results}
\setlength{\tabcolsep}{5pt}
\newcolumntype{C}{>{\centering\arraybackslash}X} 
\begin{tabularx}{\textwidth}{c l | C | C | C | C | C | C}
\toprule
\rowcolor{gray}& Method & Top-1(\%) & Top-5(\%) & Top-10(\%) & Top-20(\%) & Top-50(\%) & \smaller[0.5]Top-100(\%) \\
\cmidrule{1-8}
\addlinespace[-4pt]
\midrule
\multirow{5}{*}{\rotatebox{90}{Bamboo236}}
 & Manual Random Search & 0.80 & 4.22 & 8.40 & 17.00 & 42.30 & - \\
 & DTW~\cite{berndt1994using} & 7.20 & 22.46 & 30.93 & 49.58 & 75.42 & - \\
 & FMM~\cite{frenkel2003curve} & 8.90 & 18.64 & 26.27 & 38.14 & 65.25 & - \\
 & SIS~\cite{calabi1998differential} & 10.59 & 22.03 & 30.93 & 47.03 & 72.88 & - \\
 & \cellcolor{orange}\textbf{WisePanda} & \cellcolor{orange}\textbf{12.29} & \cellcolor{orange}\textbf{35.17} & \cellcolor{orange}\textbf{52.54} & \cellcolor{orange}\textbf{69.07} & \cellcolor{orange}\textbf{94.07} & - \\ %
\midrule
\multirow{5}{*}{\rotatebox{90}{Bamboo1350}}
 & Manual Random Search & 0.08 & 0.44 & 0.89 & 1.79 & 4.48 & 8.97 \\
 & DTW~\cite{berndt1994using} & 3.81 & 12.29 & 14.41 & 20.34 & 32.20 & 37.71 \\
 & FMM~\cite{frenkel2003curve} & 6.36 & 11.44 & 13.14 & 16.10 & 25.00 & 33.47 \\
 & SIS~\cite{calabi1998differential} & \textbf{7.20} & 15.25 & 19.07 & 24.15 & 36.02 & 48.73 \\
 & \cellcolor{orange}\textbf{WisePanda} & \cellcolor{orange}5.93 & \cellcolor{orange}\textbf{18.22} & \cellcolor{orange}\textbf{24.58} & \cellcolor{orange}\textbf{37.29} & \cellcolor{orange}\textbf{52.54} & \cellcolor{orange}\textbf{66.10} \\ %
 \midrule
\multirow{5}{*}{\rotatebox{90}{Wood}}
 & DTW~\cite{berndt1994using}[Wood670] & 2.69 & 6.72 & 11.04 & 15.82 & 23.43 & 28.51 \\
 & FMM~\cite{frenkel2003curve}[Wood670] & 2.24 & 6.72 & 9.25 & 13.28 & 21.94 & 27.61 \\
 & SIS~\cite{calabi1998differential}[Wood670] & 2.24 & 7.91 & 10.75 & 15.22 & 23.43 & 27.91 \\
 & WisePanda[Wood670] & 1.94 & 10.00 & 17.31 & 25.22 & 32.99 & 35.22 \\
 & WisePanda[Wood3833] & 1.19 & 2.84 & 5.67 & 9.25 & 15.67 & 19.85 \\ %
\bottomrule
\end{tabularx}
\begin{tablenotes}
\footnotesize
\item Results show Top-k matching accuracy (percentage of correctly identified matches within the top k candidates) across bamboo and wooden slip datasets. The original bamboo test set (Bamboo236) contains 118 paired fragments, while the expanded dataset (Bamboo1350) expands the candidate pool with 1,114 additional interference fragments. Similarly, for wooden slips, we compare performance on the original test set (Wood670) and an expanded dataset (Wood3833). As demonstrated in Extended Figure~\ref{fig:extended_robustness}, WisePanda consistently outperforms traditional curve matching approaches in both material types and across varying complexity levels, demonstrating the effectiveness of our physics-driven approach in handling diverse archaeological fragments.
\end{tablenotes}
\end{threeparttable}
\end{table}

\subsection*{Interpretability of WisePanda}
The fundamental challenge in generating training data for fragmented bamboo slip matching lies in the irreversible nature of the physical degradation process~\cite{cha2014micromorphological, li2014investigation}. When archaeologists uncover bamboo slip fragments, they observe only the final degraded curves~\cite{liu2024approaching} (Figure~\ref{fig:physics_model}c). The original fracture patterns are permanently altered by centuries of degradation and cannot be directly inferred from the preserved fragments~\cite{zhou2017identifying}. This irreversibility poses a significant limitation for conventional data-driven approaches~\cite{eslami2020review, hou2018novel, zhang2022data}, which typically require paired samples of ``before-degradation'' and ``after-degradation'' states for effective training data that is inherently unavailable in archaeological contexts. Even generative models like generative adversarial networks and diffusion models suffer from distribution bias when learning only from real fragments, failing to incorporate the underlying physical principles that govern fracture formation and degradation.

Our physics-driven model addresses this challenge by modeling the forward process of fragmentation and degradation. We first simulate the initial fracture patterns based on the physics of bamboo fiber structures~\cite{behera2025review, amada1997fiber, low2006mapping, tan2011mechanical}, then apply physical degradation processes to these patterns. This sequential modeling captures both the immediate fracture characteristics and their subsequent transformation through environmental exposure. The physical processes governing bamboo deterioration follow specific geometric principles, where protruding areas and exposed surfaces are more susceptible to degradation~\cite{cha2014micromorphological, li2014investigation}. By incorporating these physical constraints, WisePanda generates realistic degradation patterns that closely mirror natural deterioration processes, effectively bridging the gap between the unknown original fracture patterns and the observed degraded edges.

These physical principles not only enhance the quality of synthetic training data but also provide interpretability to our WisePanda's behavior. The combination of authentic fracture patterns and physically-based degradation enables WisePanda to capture fundamental characteristics of bamboo slip fragmentation, generating abundant realistic training data and leading to more robust and generalizable performance. Unlike pure data-driven approaches that may overfit to superficial patterns in limited training samples, our physics-driven model learns meaningful features grounded in the actual physical processes of fracture and degradation, allowing it to generalize effectively to unseen fragments with diverse deterioration patterns.

\subsection*{Rejoin fragmented ancient wooden slips}
To examine WisePanda's generalization capability, we applied our approach to wooden slips - another important medium for ancient text preservation that shares similar characteristics with bamboo slips but exhibits distinct material properties. While wooden slips served similar documentary purposes, their different fiber arrangement and material structure present unique challenges for fragment matching~\cite{staack2016reconstruction}. We evaluated our method on a dataset of 335 pairs of wooden slip fragments, following the same experimental protocol as with bamboo slips.

As shown in Table~\ref{tab:comparative_results}, WisePanda demonstrated promising cross-material generalization with 32.99\% Top-50 accuracy on wooden slips. When tested with extended interference fragments in both materials, performance predictably decreased to 52.54\% for bamboo slips (Bamboo1350) and 15.67\% for wooden slips (Wood3833).

This performance difference can be attributed to two key factors. First, the fiber structure of wooden slips differs substantially from bamboo slips, leading to distinct fracture patterns that deviate from our physics-driven model's assumptions~\cite{haiyan2016damage}. Second, the degradation process varies significantly due to diverse preservation conditions - factors such as burial duration, soil moisture levels, microbial activity, and regional soil composition all affect how materials deteriorate over time~\cite{singh2024pivotal}. Even artifacts of the same material excavated from different sites or time periods may exhibit varying degradation patterns, requiring different physical parameters for accurate modeling. These variations are particularly pronounced when comparing wooden and bamboo materials, as their distinct physical properties interact differently with environmental factors, leading to material-specific degradation processes that need to be carefully considered in the modeling approach.
                                                               
\subsection*{Discussion}
In this work, we present WisePanda, the first physics-driven deep learning framework designed for rejoining fragmented ancient bamboo slips. By incorporating principles from the physics of fracture into the generation of synthetic training data, our approach overcomes the fundamental challenge of paired data scarcity in AI for rejoining fragmented ancient bamboo slips, and significantly improves the efficiency of archaeologists. The effectiveness of this approach is demonstrated through superior matching accuracy and the development of practical tools that are now assisting archaeologists in their reconstruction work. Our framework's success in handling both fracture pattern generation and material degradation simulation highlights the advantage of combining physical principles with deep learning, particularly in scenarios where traditional data-driven approaches are constrained by limited training samples.

Beyond its current implementation, this research opens new horizons in bamboo slip reconstruction and broader cultural heritage preservation. For bamboo slips specifically, our framework can be extended to handle more complex scenarios by incorporating additional physical parameters that account for regional variations and different preservation conditions. Furthermore, our approach establishes a new paradigm for fragment matching where paired training data is scarce. The incorporation of physics-driven mechanism  can be adapted to other archaeological materials such as wooden slips, ceramics, metals, and pottery, by modeling their specific physics of fracture and degradation processes. 

In conclusion, WisePanda represents not merely a technical advancement in fragmented bamboo slip rejoining, but rather a transformative approach to preserving and studying cultural heritage. By combining physical principles with artificial intelligence, our method provides a robust solution when paired training data is limited but physical principles are well understood. This synergy of physics and deep learning opens new possibilities for archaeological fragment matching, potentially revolutionizing how we recover and interpret artifacts from our past.

\printbibliography[title=References - Main]
\end{refsection}

\begin{refsection}
\section*{Methods}
\subsection*{Previous work}
In recent years, various approaches have been proposed for restoring and analyzing fragmented ancient artifacts~\cite{zhou2017identifying, zhang2018multi}. Traditional methods have primarily focused on geometric feature analysis, particularly curve matching methods. Modern approaches mainly leverage artificial intelligence techniques~\cite{das2022role, tsatsanashvili2024artificial}. Notable recent work includes the restoration of Pompeii's frescoes using a robotic system for automated fragment matching~\cite{tsesmelis2024re}, Ithaca, a deep neural network system that successfully restored and attributed ancient Greek texts~\cite{assael2022restoring}, and the reunion helper for Dunhuang manuscript fragments~\cite{zheng2024reunion, zhang2024llmco4mr}. For oracle bone fragments, researchers have explored using GAN to generate training samples for restoration tasks~\cite{li2024oracle}.

The manual restoration of bamboo and wooden slips has a long history in China~\cite{staack2023bindings, chang1964written, needham1983science}. Wang Guowei's pioneering work \textit{Liusha Zhuijian}~\cite{LuoWang1914} systematically documented traditional techniques for matching and rejoining ancient bamboo and wooden manuscripts, establishing foundational methodologies that are still valuable today. However, manual restoration remains an extremely time-consuming and labor-intensive task, often requiring weeks or months to successfully match a single pair of fragments. With recent technological advancements, computational approaches and artificial intelligence have emerged as powerful tools to assist archaeologists in this complex process, significantly accelerating the restoration workflow while maintaining the essential role of expert knowledge. While deep learning approaches have shown promise in cultural relic restoration, the application to fragmented bamboo slips faces significant challenges due to the lack of sufficient paired training data and the complex physical degradation patterns unique to bamboo materials. Our work addresses these limitations by introducing physics-driven data generation. Unlike previous approaches, we uniquely integrate fracture mechanics and material degradation principles into the restoration process. This novel approach allows us to generate physically plausible paired training data that reflects the actual physical processes of how bamboo slips break and degrade over time.

Moreover, our physics-driven approach not only provides interpretable results that align with archaeological domain knowledge, but also demonstrates strong extensibility to various archaeological fragment restoration tasks. The framework we propose may open a new paradigm for applying deep learning in archaeological restoration~\cite{ghaith2024qualitative}, particularly valuable when paired training data is scarce but physical principles are well understood. This methodology can potentially be adapted to restore other types of archaeological fragments by modeling their specific physical properties and degradation processes~\cite{aldakheel2025physics, lesar2013introduction}, offering a generalizable solution for the broader field of cultural heritage preservation.

\subsection*{Physical model for data generation}
\noindent\textbf{Fracture modeling.} Bamboo slip fragments exhibit two distinct fracture modes: transverse and longitudinal, with transverse fractures accounting for approximately 70\% of cases~\cite{ramful2022investigation}. Due to bamboo's unique fiber structure, the transverse fractures occur when stress cuts across the bamboo slip's cross-section, creating uneven, wavy edge curves. The longitudinal fractures happen along the fiber direction, resulting in straight edges, as illustrated in Extended Figure~\ref{fig:bamboo_slips_fractures}. In this work, we focus on transverse fractures due to their prevalence and complexity. To systematically model the bamboo slip structure, we establish a coordinate system where the $x$-axis represents the horizontal direction along the bamboo slip's width and the $y$-axis represents the vertical direction corresponding to fiber bundle heights. We abstract the bamboo slip as a series of $N_f$ vertically aligned fiber bundles, as shown in Extended Figure~\ref{fig:physical_model}a. Each fiber bundle is represented as a discrete structural unit with a uniform width of $\delta_x$, forming a continuous arrangement along the horizontal axis. The initial height for the front of $i$-th fiber bundle before environmental degradation is denoted as $y_i^{init}$. The final height after corrosion processes is represented as $y_i^{final}$. This discretization allows us to precisely model the propagation of fracture across the material's anisotropic structure. The vertical positions of these fiber bundles determine the morphology of the fracture curve, with the height of each bundle representing its position on the fracture curve. When modeling the fracture process, we observe that bamboo slips were typically buried in a loosely rolled form, causing them to experience predominantly Mode III (out-of-plane shear) stress during burial compression and subsequent deterioration. This specific stress mode occurs when opposing forces act parallel to each other but in different planes, creating a tearing effect particularly relevant to rolled manuscripts. We model the bamboo slip structure as consecutive fiber bundles~\cite{dixon2014structure, chen2019fracture, ramful2020investigation, huang2017comparison}, where stress propagates from one bundle to another one (Figure~\ref{fig:physics_model}a). At the terminal point of each fractured fiber, a stress component $K_{\mathrm{III}}$ continues to propagate~\cite{gdoutos2020fracture}, following the formula:
\begin{equation}
    K_{\mathrm{III}}=\sqrt{2\pi}\sigma\sqrt{\alpha},
\end{equation}
where $\sigma$ represents the far-field shear force and $\alpha$ denotes the cumulative stress extension path length. The stress field at the fracture propagation front $\sigma_{\mathrm{yz}}(r,\theta)$ can be described by:
\begin{equation}
    \sigma_{\mathrm{yz}}(r,\theta)=\frac{K_{\mathrm{III}}}{\sqrt{2\pi r}}\cos\left(\frac{\theta}{2}\right),
\end{equation}
where $r$ is the radial distance from the fracture propagation front and $\theta$ is the angular position relative to the direction of stress propagation, as shown in Extended Figure~\ref{fig:physical_model}b. This equation describes how the stress force decreases with distance from the fracture propagation front while also varying with direction.

For the $xy$ plane of the writing surface, we develop a probabilistic model to describe how fracture propagates across consecutive fiber bundles. As illustrated in Extended Figure~\ref{fig:physical_model}b, when a fracture reaches point \textit{P$_2$} at the end of fiber bundle $f_{i-1}$, it generates a stress field that determines the probable direction of continued fracture propagation into the next fiber bundle $f_i$. The blue dotted circles emanating from point \textit{P$_2$} represent this stress field distribution, with point \textit{P$_1$} marking the beginning of the fracture on bundle $f_{i-1}$ and point \textit{P$_3$} indicating the potential endpoint of the fracture on bundle $f_i$. The stress field on the $xy$ plane is described by:
\begin{equation}
\sigma_{\mathrm{xy}}(r,\theta) = V_{\mathrm{xy}} \cdot \frac{K_{\mathrm{III}}}{\sqrt{2\pi r}}\cos\left(\frac{\theta}{2}\right),
\end{equation}
where $V_{\mathrm{xy}}$ is a dimensionless constant representing the projection of out-of-plane (Mode III) stress onto the writing surface plane, $r$ is the radial distance from the fracture propagation front at point \textit{P$_2$}, and $\theta$ is the angular position in the stress field. This formulation accounts for bamboo's anisotropic fiber structure in translating three-dimensional stress into planar fracture propagation. The angles $\theta_{i-1}$ and $\theta_i$, measured relative to the horizontal axis, define the fracture directions in consecutive fiber bundles, with $\theta_i = \theta_{i-1} + \Delta\theta$. The stress field determines the probability distribution of angle change $\Delta\theta$, which governs how the fracture direction transitions between fiber bundles. This probability density function can be expressed as:
\begin{equation}
p(\Delta\theta|\theta_{i-1}) = \frac{V_{\mathrm{xy}} \cdot K_{\mathrm{III}} \cdot \cos\left(\frac{\Delta\theta}{2}\right)}{\sqrt{2\pi}(\delta_x\cos(\theta_{i-1}+\Delta\theta))}.
\end{equation}
By substituting $K_{\mathrm{III}} = \sqrt{2\pi}\sigma\sqrt{\alpha}$, we simplify the above equation to:
\begin{equation}
p(\Delta\theta|\theta_{i-1}) = \frac{V_{\mathrm{xy}} \cdot \sigma\sqrt{\alpha} \cdot \cos\left(\frac{\Delta\theta}{2}\right)}{\delta_x\cos(\theta_{i-1}+\Delta\theta)}.
\label{eq:five}
\end{equation}
While $V_{\mathrm{xy}}$ and $\sigma$ are unmeasurable constants that scale the probability distribution, they do not affect the relative probabilities of different fracture paths. Once the fracture angle is determined, the initial vertical height relationship between adjacent fiber bundles follows:
\begin{equation}
y_{i+1}^{init} - y_i^{init} = \tan(\theta_i) \cdot \delta_x.
\label{eq:six}
\end{equation}
This mathematical modeling enables us to construct a Markov chain-like model for fracture propagation, where each state represents a fiber bundle's initial front height $y_i^{init}$, and transitions between states are governed by the probability density function $p(\Delta\theta|\theta_{i-1})$. This results in a physically plausible model that captures how Mode III stress-induced fracture propagates through bamboo's anisotropic structure, allowing us to generate diverse yet authentic synthetic fracture patterns that serve as the foundation for subsequent corrosion simulation to obtain the final degraded fracture curves.

\medskip
\noindent\textbf{Corrosion process simulation.} 
After the initial fracture curves are generated, the bamboo slip fragments undergo long-term environmental degradation that significantly alters their original morphology. The degradation process of bamboo slip fragments is highly dependent on their geometric features, where sharp edges and protrusions are more susceptible to corrosion compared to smooth surfaces~\cite{forest1987wood, blanchette1994assessment}, even with identical perimeter lengths. This geometric dependency is primarily due to stress concentration at sharp points and increased surface area exposure to environmental factors~\cite{forest1987wood}. We observe that in archaeological samples, protruding areas of bamboo slip edges show significantly more deterioration than recessed areas, consistent with their greater environmental exposure. To mathematically capture this geometric dependency in our corrosion model, we employ the $\mathrm{ReLU}$ function to calculate the structural exposure area for each fiber bundle. We define the structural exposure area $S_i$ as:
\begin{equation}
S_i = \mathrm{ReLU}(y_i^{init} - y_{i-1}^{init}) + \mathrm{ReLU}(y_i^{init} - y_{i+1}^{init}).
\label{eq:seven}
\end{equation}
%
The erosion process is then modeled by reducing the fiber height based on the structural exposure area and a corrosion rate coefficient $c_{\mathrm{e}}$. The final degraded height for the front of $i$-th fiber bundle $y_i^{final}$ is given by:
\begin{equation}
y_i^{final} = y_i^{init} - (S_i \times c_{\mathrm{e}}).
\label{eq:eight}
\end{equation}
%
This approach ensures that regions with higher geometric exposure experience accelerated degradation, accurately simulating how moisture and microorganisms propagate through the bamboo structure~\cite{forest1987wood, blanchette1994assessment, liu2025cracks}. To capture the progressive nature of long-term degradation, we implement an iterative corrosion process where the degradation is applied over multiple time steps, with each iteration representing a period of environmental exposure. The synchronized update mechanism ensures that all fiber heights are calculated based on the current state before updating to the next iteration, preventing cascade effects where early updates could disproportionately influence later calculations within the same corrosion cycle. This multi-step simulation accurately reflects the physical degradation process observed in actual bamboo slips, where environmental factors affect the entire surface concurrently rather than sequentially, ultimately generating realistic synthetic fragment pairs that capture both the initial fracture characteristics and their subsequent transformation through centuries of environmental exposure.

\medskip
\noindent\textbf{Parameter optimization.}
To ensure our physical model accurately reflects real bamboo slip characteristics, we develop a parameter optimization strategy driven by comparison between real and synthetic data distributions. We first select a set of 200 real fragment curves (set A) extracted from archaeological bamboo slips, which represent the final state after both fracture formation and corrosion processes, with curves characterized by their $y^{final}$ coordinates along the fragment curve. Concurrently, we generate a set of 200 simulated fragment curves (set B) using randomly initialized parameters in our physical model, where the synthetic fragments undergo the complete process from initial fracture generation ($y^{init}$ heights) through corrosion simulation to obtain final degraded curve profiles ($y^{final}$ heights). Both sets are subjected to t-SNE (t-Distributed Stochastic Neighbor Embedding)~\cite{van2008visualizing} dimensionality reduction, projecting the high-dimensional curve data into a 2D space to facilitate comparison of their distributions. The silhouette score $s$ for these reduced data sets is calculated as:
\begin{equation}
s(j) = \frac{b(j) - a(j)}{\max\{a(j), b(j)\}},
\label{eq:nine}
\end{equation}
where $a(j)$ represents the mean distance between point $j$ and all other points within its own set (either set of real fragments A or set of synthetic fragments B), and $b(j)$ is the mean distance between point $j$ and all points in the other set. A silhouette score approaching zero indicates that the boundary between real and synthetic data distributions becomes increasingly indistinguishable, suggesting optimal similarity between our simulated fragmented curves and actual archaeological fragments. To find the optimal set of parameters for our physical model, we employ a genetic algorithm~\cite{Goldberg1989} with the fitness function $F$ as:
\begin{equation}
F = \frac{1}{|s|} + \lambda \cdot \mathrm{diversity},
\label{eq:ten}
\end{equation}
where $\lambda$ is a weight coefficient that balances optimization precision (minimizing silhouette score) with exploration capability (maintaining genetic diversity). The genetic algorithm manages a population of parameter configurations $\mathcal{P} = \{p_1, \ldots, p_n\}$, where each parameter set $p_i$ contains specific values for both fracture physics and corrosion simulation components of our model. During each optimization iteration, new candidate parameter sets are generated through controlled recombination:
\begin{equation}
p_{\mathrm{new}} = \beta \cdot p_1 + (1-\beta) \cdot p_2 + \epsilon,
\label{eq:eleven}
\end{equation}
where $p_1$ and $p_2$ are parent parameter sets selected based on their fitness scores, $\beta \in [0,1]$ is a randomly generated crossover weight that varies with each recombination operation, and $\epsilon$ represents a small random mutation noise term added to maintain genetic diversity. This randomized crossover mechanism allows the algorithm to explore different combinations of parameter values while gradually converging towards configurations that produce simulated fracture patterns closely matching the real archaeological data~\cite{Goldberg1989}.

\subsection*{Data preparation and augmentation}
\noindent\textbf{Fracture curve generation.} Based on our analysis of Mode III (out-of-plane shear) stress-induced fracture mechanisms in bamboo slips~\cite{anderson2005fracture}, which typically occurs when rolled bamboo manuscripts experience perpendicular forces during burial and subsequent deterioration, we implement a probabilistic framework to generate initial fracture patterns. The generation process leverages Eq.~\eqref{eq:five}, Eq.~\eqref{eq:six}, and Eq.~\eqref{eq:seven} to construct a Markov chain-like structure, where each state represents the initial height $y_i^{init}$ of $i$-th fiber bundle (Algorithm \ref{alg:fracture_generation}). This algorithm generates fracture curves that closely mimic the physical behavior of bamboo fiber bundles under stress. The probabilistic nature of the generation process, governed by the Probability Density Function (PDF) derived from stress field analysis in Eq.~\eqref{eq:five}, ensures that each generated curve exhibits realistic variations while maintaining the fundamental characteristics of bamboo slip fracture. These generated curves serve as the foundation for subsequent corrosion simulation.
%
\begin{small}
\begin{algorithm}[H]
    \caption{Physics-Driven Fracture Pattern Generation}
    \label{alg:fracture_generation}
    \SetKwInOut{Input}{Input}
    \SetKwInOut{Output}{Output}
    \SetKwFunction{FSample}{SampleAngleChange}

    \Input{
        $\delta_x$: Width of each fiber bundle \\
        $N_f$: Number of vertically aligned fiber bundles \\
        $\theta_{init}$: Initial fracture angle at $(x_0, y_0) = (0, 0)$ \\
    }
    \Output{
        $P^{init} = [(x_0, y_0^{init}), (x_1, y_1^{init}), \ldots]$
    }

    \BlankLine
    \tcp{Initialization}
    $P^{init} \leftarrow \text{new List}()$ \\
    $current\_x \leftarrow 0.0$ \\
    $current\_y^{init} \leftarrow 0.0$ \\
    $current\_\theta \leftarrow \theta_{init}$ \\
    Append $(x_0 = current\_x, y_0^{init} = current\_y^{init})$ to $P^{init}$ \\
    $i \leftarrow 0$
    
    \BlankLine
    \tcp{Iterative generation across the width}




     \While{$i < N_f$}{
        \tcp{Sample the change in angle based on Eq.~\eqref{eq:five} }
        $\Delta\theta \leftarrow \FSample(current\_\theta)$ \\

        \tcp{Calculate next state}
        $next\_\theta \leftarrow current\_\theta + \Delta\theta$ \\
        $\Delta y^{init} \leftarrow \tan(next\_\theta) \times \delta_x$ \\
        $next\_y^{init} \leftarrow current\_y^{init} + \Delta y^{init}$ \\
        $next\_x \leftarrow (i + 1) \times \delta_x$ \\

        \tcp{Update current state for next iteration}
        $current\_x \leftarrow next\_x$ \\
        $current\_y^{init} \leftarrow next\_y^{init}$ \\
        $current\_\theta \leftarrow next\_\theta$ \\
        $i \leftarrow i + 1$ \\

        \tcp{Record the new point}
        Append $(x_i = current\_x, y_i^{init} = current\_y^{init})$ to $P^{init}$ \\
    }

    \BlankLine
    \Return{$P^{init}$}
    \label{alg:alg1}
\end{algorithm}
\end{small}

\medskip
\noindent\textbf{Corrosion simulation.} 
Building upon Eq.~\eqref{eq:six}, Eq.~\eqref{eq:seven} and Eq.~\eqref{eq:eight}, we developed a multi-scale corrosion model that quantifies degradation based on the geometric exposure of individual fibers and their spatial relationships with adjacent fibers. The $\mathrm{ReLU}$ function effectively captures only the protruding portions that are more vulnerable to environmental degradation (Algorithm \ref{alg:corrosion_simulation}). Algorithm \ref{alg:corrosion_simulation} implements a synchronized update mechanism where all fiber heights' expected changes are calculated first, and then all heights are updated simultaneously at each time step. This approach prevents cascade effects where early fiber updates could disproportionately influence later calculations within the same corrosion cycle. The number of iterations $N_c$ and corrosion rate $c_e$ are determined through parameter optimization to match observed degradation patterns in archaeological samples. This two-phase process - height calculation followed by simultaneous update - ensures that the corrosion simulation accurately reflects the physical degradation process observed in actual fragmented bamboo slips, where environmental factors affect the entire surface concurrently rather than sequentially. The algorithm transforms the initial fracture curves into realistic degraded fragments, completing the synthetic data generation pipeline.
\begin{small}
\begin{algorithm}[H]
    \caption{Multi-Step Corrosion Simulation}
    \label{alg:corrosion_simulation}
    \SetKwInOut{Input}{Input}
    \SetKwInOut{Output}{Output}
    \SetKw{KwCopy}{copy}
    
    \Input{
        $Y^{init}$: Array of initial fiber heights $[y_1^{init}, y_2^{init}, \ldots, y_{N_f}^{init}]$ \\
        $N$: Number of corrosion simulation iterations \\
        $N_f$: Number of vertically aligned fiber bundles \\
        $c_e$: Corrosion rate coefficient
    }
    \Output{
        $Y^{final}$: Array of fiber bundle heights after $N$ iterations $[y_1^{final}, y_2^{final}, \ldots, y_{N_f}^{final}]$
    }

    \BlankLine
    \tcp{Initialization}
    $Y^{current} \leftarrow \KwCopy(Y^{init})$ \tcp{Start with initial heights}
    $N_f \leftarrow \text{length}(Y^{current})$ \tcp{Number of fiber bundles}

    \BlankLine
    \tcp{Perform $N$ iterations of corrosion}
    \For{iteration from $1$ to $N$}{
        $Y^{next} \leftarrow \text{new Array(} N_f \text{)}$

        \For{$i$ from $1$ to $N_f$}{
            $y_{curr} \leftarrow Y^{current}[i]$ 

            \eIf{$i > 1$}{
                $y_{prev} \leftarrow Y^{current}[i-1]$
            }{
                $y_{prev} \leftarrow y_{curr}$ 
            }
            \eIf{$i < N_f$}{
                $y_{next} \leftarrow Y^{current}[i+1]$
            }{
                $y_{next} \leftarrow y_{curr}$ 
            }

            \tcp{Calculate structural exposure using $\mathrm{ReLU}(x) = \max(0, x)$}
            $exposure \leftarrow \max(0, y_{curr} - y_{prev}) + \max(0, y_{curr} - y_{next})$ \\

            \tcp{Calculate height after erosion}
            $eroded\_height \leftarrow y_{curr} - (exposure \times c_e)$ \\

            \tcp{Store result for the next state}
            $Y^{next}[i] \leftarrow eroded\_height$ \\
        }

        \tcp{Update the current state simultaneously for the next iteration}
        $Y^{current} \leftarrow Y^{next}$ \\
    }

    \BlankLine
    $Y^{final} \leftarrow Y^{current}$ \\
    \Return{$Y^{final}$}
    \label{alg:alg2}
\end{algorithm}
\end{small}

\medskip
\noindent\textbf{Parameter tuning through genetic algorithm.} To ensure our physical model accurately reflects real bamboo slip characteristics, we developed a parameter optimization strategy driven by comparison between real and synthetic data distributions. The parameters in our physical model have physical interpretations that can be loosely connected to real-world properties~\cite{hussain2024machine, kutz2023machine}, such as bamboo fiber structure, material characteristics, and environmental degradation factors. While these parameters are not direct measurements of physical properties, they collectively contribute to modeling the complex fracture and corrosion processes observed in archaeological bamboo slip fragments~\cite{brynjarsdottir2014learning, matthiesen2015detecting}. We collect curves from real fragmented bamboo slips, which represent the final state after both fracture and corrosion processes, characterized by their $y^{final}$ coordinates along the fragment boundary. These curves undergo dimensionality reduction using t-SNE to create a reference distribution. Concurrently, we generate synthetic data using randomly initialized parameters and apply the same dimensionality reduction process. Our synthetic fragments undergo the complete physical simulation pipeline: starting with initial fracture generation to obtain $y^{init}$ coordinates, followed by corrosion simulation to produce final degraded profiles with $y^{final}$ coordinates that can be directly compared with real archaeological data. The optimization objective is to minimize the distributional difference between real and synthetic data, quantified by the silhouette coefficient in Eq.~\eqref{eq:nine}. Our genetic algorithm iteratively refines the parameter set by maintaining a population of parameter configurations $\mathcal{P} = \{p_1, \ldots, p_{n}\}$, where each $p_i$ contains parameters for both fracture and corrosion models~\cite{Goldberg1989}. The visualization in Extended Figure~\ref{fig:extended_data_parameter_optimization} confirms that optimizing these physical parameters directly improves the model's ability to identify correct fragment matches. The fitness function follows Eq.~\eqref{eq:ten}, which combines the silhouette score with a diversity term for balancing between distribution matching and population diversity. New parameter sets are generated through Eq.~\eqref{eq:eleven}, where $p_1$ and $p_2$ are selected based on fitness scores, $\beta$ is a random crossover weight, and $\epsilon$ represents mutation noise. The effectiveness of our parameter optimization approach is visually demonstrated in Extended Figure~\ref{fig:extended_data_parameter_optimization}, where the distribution of synthetic data (blue points) progressively aligns with the real data distribution (green points) through iterations. The corresponding heatmap of matching quality shows stronger correlation patterns as the silhouette coefficient improves, validating our parameter optimization strategy~\cite{rousseeuw1987silhouettes}. This optimization ensures that our synthetic fragments accurately capture both the initial fracture characteristics and the subsequent degradation patterns, enabling the deep learning model to learn features grounded in actual physical processes.

\subsection*{WisePanda architecture}
\noindent\textbf{Input representation.} Each input to WisePanda consists of bamboo slip fragment curve data encoded as 64-dimensional vectors. We extract these vectors through a straightforward sampling process: after detecting and isolating the fracture curve from each fragment image, we normalize the curve length and sample 64 equidistant points along the curve from left to right. The resulting sequence of 64 heights capture the distinctive geometric profile of each fracture curve while providing a standardized input format for the following neural network. This fixed-dimensional representation ensures consistent processing across fragments of different sizes and complexities while preserving the key morphological features that distinguish potential matching pairs.

\medskip
\noindent\textbf{Network design.} The core of WisePanda's architecture is a TripletNet-based deep learning network designed for similarity learning between fragment pairs. The network processes triplets of fragments: an anchor sample $c_a$ (a fragmented curve), a positive sample $c_p$ (a matching fragment), and a negative sample $c_n$ (a non-matching fragment). Each branch of the TripletNet shares identical weights to maintain consistency in feature extraction, and comprises three key components: a local feature encoder that captures fine-grained geometric patterns using 1D convolutions, a self-attention mechanism for modeling long-range dependencies in the curve, and a cross-fragment attention module that learns to align corresponding patterns between potential matches~\cite{hoffer2015deep}. This network is trained using a triplet loss function:
\begin{equation}
\mathcal{L}_{\mathrm{triplet}} = \max(0, d(c_a, c_p) - d(c_a, c_n) + m),
\end{equation}
where $d(c_a, c_p)$ represents the distance between anchor and positive samples in the learned feature space, $d(c_a, c_n)$ is the distance between anchor and negative samples, and $m$ is the margin parameter that enforces a minimum separation between matching and non-matching pairs. This loss function effectively guides the network to learn a feature space where matching fragments are pulled closer together while pushing non-matching fragments apart~\cite{hoffer2015deep}. The network outputs a similarity score within [0,1] for each potential match, where values closer to 1 indicate higher match probability. 


\medskip
\noindent\textbf{Ranking mechanism.} To facilitate archaeological fragment rejoining, WisePanda implements a ranking-based matching system that provides archaeologists with a manageable set of potential matches~\cite{funkhouser2011learning, di2022review}. For each input fragment, the network computes similarity scores with all candidate fragments in the dataset. These scores are then sorted to generate a ranked list of the Top-$k$ most probable matches, typically $k=50$, significantly reducing the search space from thousands of possibilities to a practical subset for further manual verification. The ranking process is guided by the predicted similarity metric from our TripletNet network architecture. When a new fragment is presented, the system: 1) extracts the fracture curve and generates the 64-dimensional feature vector; 2) computes pair-wise similarity scores with existing fragments through the trained network; 3) ranks all candidates based on their similarity scores; and 4) returns the Top-$k$ candidates with the highest matching probabilities. This approach transforms WisePanda from a theoretical matching model into a practical archaeological tool that effectively guides the rejoining process while involving human expertise in the final verification steps.

\subsection*{Training details}
WisePanda was trained using two NVIDIA RTX 4090 GPUs with a batch size of 100. The network was trained using the Adam optimizer with an initial learning rate of 1e-3 and a one-cycle learning rate scheduler~\cite{smith2019super}. 
We implemented a combined loss function that extends the concept of triplet loss to a more direct optimization objective:
\begin{equation}
\mathcal{L} = E[(d_{\mathrm{pos}} - 0)^2 + (d_{\mathrm{neg}} - 1)^2]
\end{equation}
where $d_{\mathrm{pos}}$ represents the distance between matching pairs (which we want to push toward 0) and $d_{\mathrm{neg}}$ represents the distance between non-matching pairs (which we want to push toward 1). This squared error formulation provides smoother gradients compared to the standard triplet loss while maintaining the same conceptual goal of maximizing the separation between matching and non-matching fragments. We utilized PReLU activation functions for non-linear transformations, which adapt to the data by learning the optimal negative slope parameter rather than using a fixed value as in standard ReLU. 

The input fragmented curves were normalized by rescaling both $x$ and $y^{final}$ coordinates to fall within the [0, 1] range before being sampled and fed into the network. The training process ran for 150 epochs to ensure convergence, with model parameters being updated approximately 100,000 times. To ensure reproducibility, model training was conducted with fixed random seeds. For inference, we employ an efficient ranking strategy where each fragment is compared with the entire database using the trained similarity metric. The matches are then sorted by matching probabilities to generate the Top-$k$ recommendations for archaeological verification.

\subsection*{Baseline methods}
To evaluate WisePanda's effectiveness, we compare it against three traditional methods widely used in fragment matching and curve alignment tasks. 

\medskip
\noindent\textbf{Traditional curve matching baseline.} The traditional curve matching approach employs dynamic time warping (DTW)~\cite{berndt1994using} to measure the geometric similarity between fragmented curves. This method directly compares the normalized $(x, y^{final})$ coordinate sequences of two fragments, characterizing their alignment cost as a measure of potential matching.

\medskip
\noindent\textbf{Scale Invariant Signature (SIS).} SIS~\cite{calabi1998differential} transforms fragmented curves into scale-invariant representations by computing local curvature features along the fragmented curve. This transformation generates signatures that are independent of fragment size and orientation, enabling comparison through simple distance metrics.

\medskip
\noindent\textbf{Fast Matching Method (FMM).} FMM~\cite{frenkel2003curve} accelerates the matching process by decomposing fragmented curves into key geometric features and employing a hierarchical matching strategy. It first identifies potential matches using coarse features, then refines the results using more detailed geometric information.

\subsection*{Evaluation metrics}
\noindent\textbf{Top-k accuracy.} To assess WisePanda's performance in bamboo slip rejoining, we employ the Top-$k$ accuracy metric, which measures the percentage of correctly identified matches within the $k$ highest-ranked suggestions. For each query fragment, a match is considered successful if the true matching fragment appears among the Top-$k$ candidates proposed by the model. We report results for $k \in {1, 5, 10, 20, 50, 100}$, providing a comprehensive view of the model's ranking capability at different thresholds of consideration. This metric directly reflects the system's practical utility in archaeological workflows, where experts typically examine a limited number of the most probable matches.

\subsection*{Future work}
While our current work demonstrates the effectiveness of physics-driven deep learning for fragmented bamboo slip rejoining, several promising directions remain for future exploration. First, our present approach primarily addresses transverse fractures, which account for approximately 70\% of bamboo slip breakages. Extending WisePanda to handle longitudinal fractures (Figure~\ref{fig:bamboo_slips_fractures})—which occur along the fiber direction and produce distinctly different breakage patterns—represents an important next step. For these longitudinal fractures, where the straight physical breakage features are less distinctive, focusing on text continuity and character alignment would be more effective than attempting to model the physical fracture properties. This would entail developing complementary methods that leverage optical character recognition and textual coherence to guide the matching process when geometric features alone are insufficient.

Beyond bamboo slips, the methodology established in this work opens avenues for broader applications in archaeological artifact restoration. Particularly promising is the extension to pottery and ceramic fragment matching, where material properties and degradation processes differ significantly from bamboo. Ceramics exhibit distinctive fracture mechanics governed by brittleness, firing conditions, and microstructure, which could be modeled using adaptations of our physics-driven approach. The translational potential of our framework to these domains could significantly advance digital archaeology tools across diverse cultural heritage contexts.

From a methodological perspective, integrating expert knowledge more deeply into the learning process presents another promising direction. Future systems might implement active learning paradigms where archaeologist feedback continuously refines the rejoining models, creating a virtuous cycle of human-AI collaboration. Additionally, exploring self-supervised or few-shot learning approaches could further reduce the dependence on synthetic data, potentially capturing even more nuanced aspects of artifact deterioration that are challenging to model explicitly.

These future directions share a common goal: bridging the gap between physical understanding and computational methods in cultural heritage preservation. By continuing to combine domain-specific physical principles with advanced machine learning techniques, we envision creating increasingly powerful tools that amplify rather than replace human expertise in the vital work of preserving and understanding our shared cultural past.


\subsection*{Competing interests}
The authors declare no competing interests.

\subsection*{Author contributions}
J.Z. and Y.X. conceived the study. H.L., X.W., J.L.L. and J.L. provided archaeological data and domain knowledge. J.Z. and Z.Z performed experiments, analyzed data, wrote the code and manuscript. J.Z., Q.T. and J.S. contributed to the literature review and software implementation. G.X. participated in methodology discussions and manuscript organization. B.D. participated in methodology discussions, data analysis, and manuscript editing. Y.X. participated in methodology discussions, experimental design, manuscript organization and revision.

\subsection*{Data availability}
The datasets used for training and evaluation are publicly available at GitHub \url{https://github.com/zhujinchi/wisepanda} under Apache License 2.0. The repository includes fragment matching datasets: Bamboo236 (118 paired bamboo fragments), Wood670 (335 paired wooden fragments), and their extended versions Bamboo1350 and Wood3833 with additional 1,114 and 3,163 interference fragments respectively. Due to confidentiality requirements for unpublished archaeological materials, the original high-resolution fragment images cannot be released. Instead, the repository provides fracture curves, cropped edge images, and synthetic training data generated by our physics-driven model. The complete data processing workflow for extracting fracture curves from archaeological fragments and generating synthetic paired data is included in the codebase. All morphological features and curve data necessary for reproducing the fragment matching experiments are available in the repository.

\subsection*{Code availability}
The source code for WisePanda framework is publicly available on GitHub at \url{https://github.com/zhujinchi/wisepanda} under Apache License 2.0. The repository contains the implementation of our physics-driven deep learning model for bamboo slip fragment rejoining, including the fracture modeling algorithms, corrosion simulation processes, neural network architecture, and user interface. Users can install the required dependencies by following the installation instructions provided in the repository. The codebase is designed to be accessible to researchers, featuring an intuitive graphical user interface for fragment selection, comparison, and verification with integrated AI assistance. Additionally, we have established a project website at \url{https://wisepanda.tech/} that provides demonstration videos, visual examples, and detailed documentation of the fragment rejoining and visualization tools discussed in this paper. This open-source release aims to facilitate further research in digital archaeology and enable the application of our methods to other types of archaeological fragment rejoining challenges.

\printbibliography[title=References - Methods]
\end{refsection}

\section*{Extended Figure}
\let\originalthefigure\thefigure

\renewcommand{\thefigure}{E1}
\begin{figure}[H]
\centering
\includegraphics[width=1\linewidth]{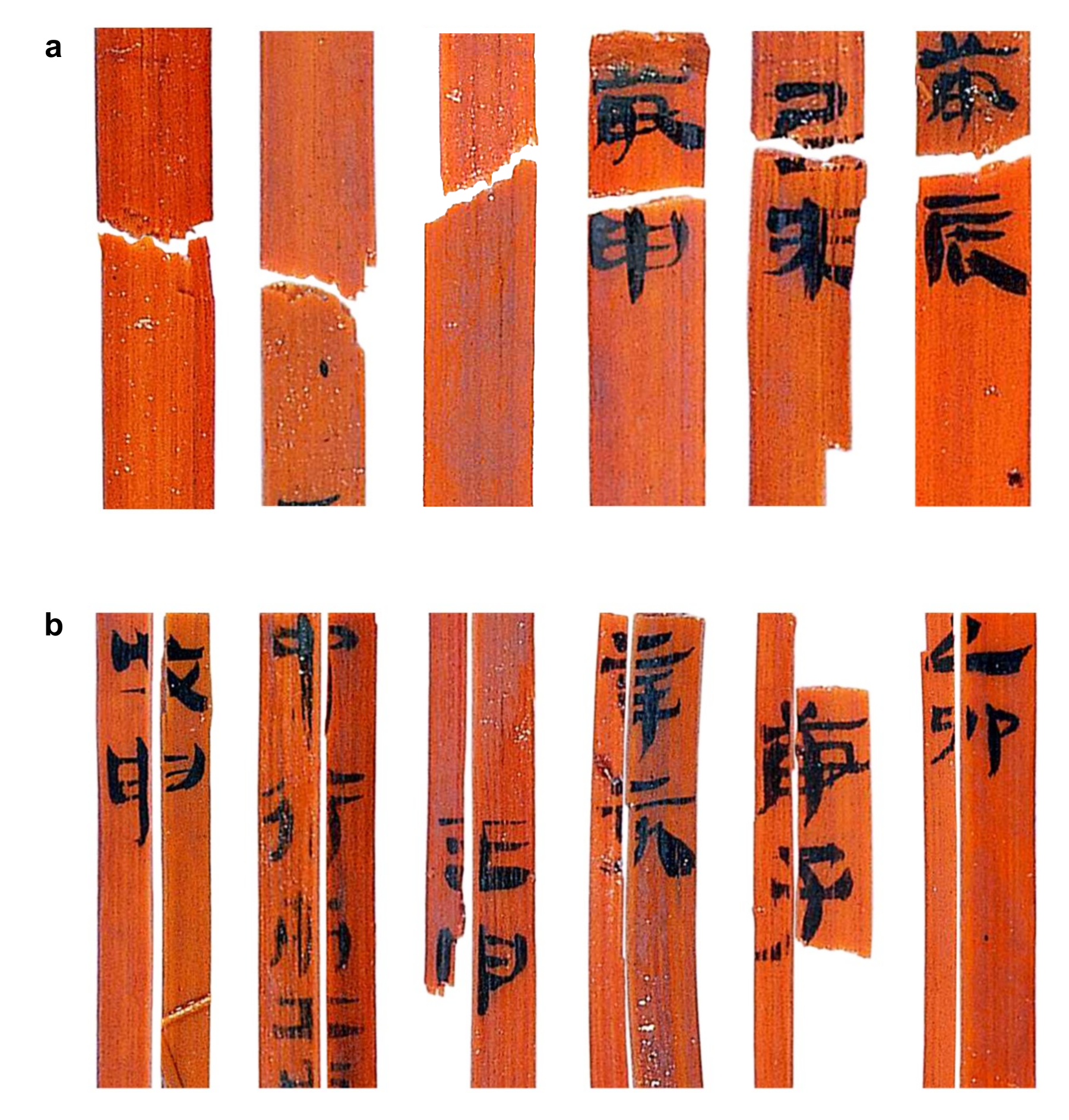}
\caption{\textbf{Typical fracture patterns of ancient bamboo slips.} (\textbf{a}), The upper row shows bamboo samples with transverse fractures, the most common fracture mode (approximately 70\%), characterized by irregular wavy breakage patterns. Most transverse fracture curves contain no text, with only occasional fragments showing partial characters, making morphological features the primary basis for rejoining. (\textbf{b}), The lower row exhibits longitudinal fractures, characterized by relatively smooth fracture curves along the fiber direction. These fragments typically preserve complete text, making textual continuity the primary criterion for matching. This diversity of fracture patterns and text preservation highlights the complexity of ancient bamboo slip rejoining.}
\label{fig:bamboo_slips_fractures}
\end{figure}
\vfill

\clearpage
\renewcommand{\thefigure}{E2}
\begin{figure}[H]
\includegraphics[width=1\linewidth]{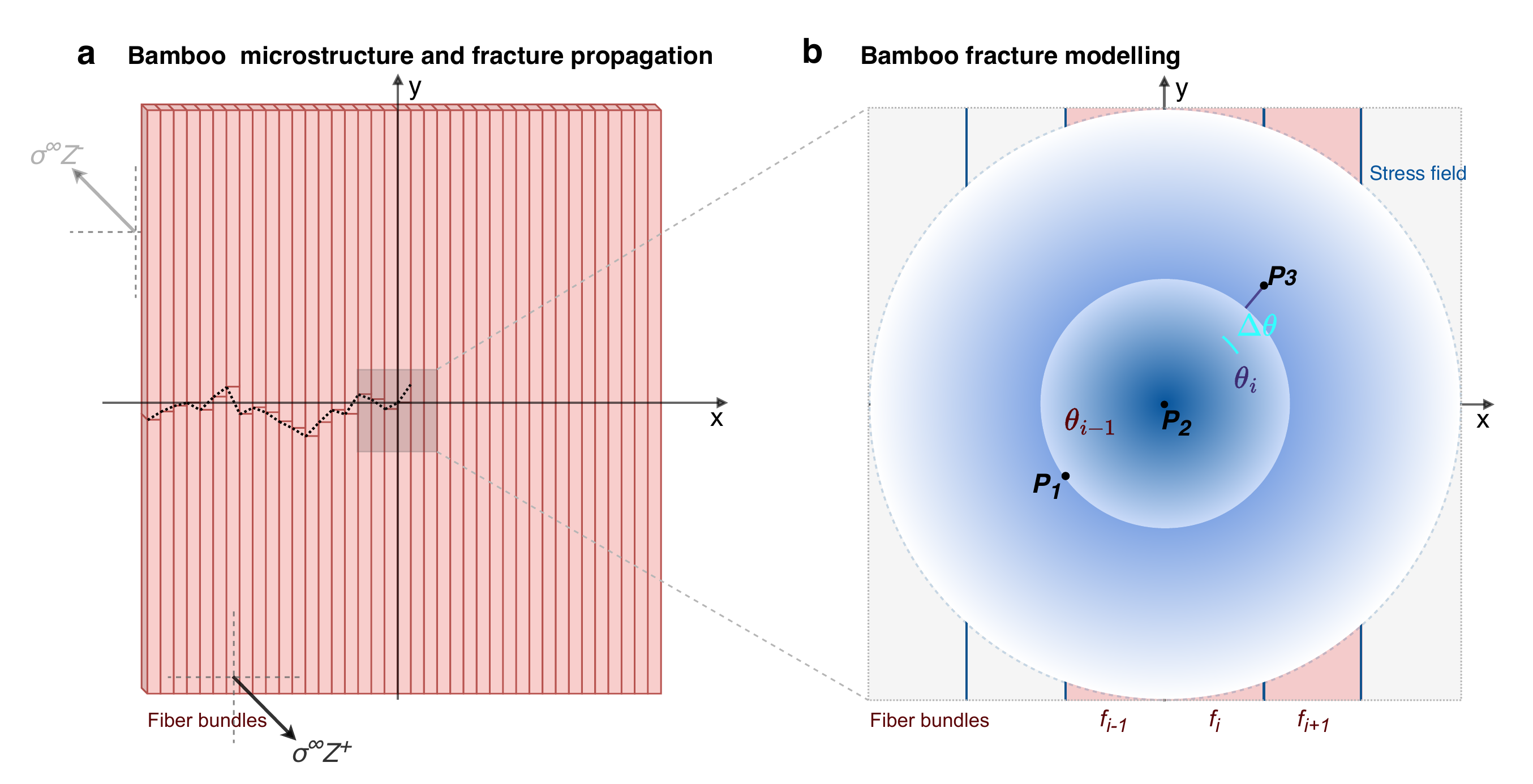}
\caption{\textbf{Physics-driven model of bamboo fracture propagation.} \textbf{a}, Bamboo microstructure showing vertical fiber bundles (red blocks) and a possible fracture path (composed of dotted black segments) traversing across fibers. The black arrows indicate shear forces ($\sigma^{\infty}Z^{+}$) inducing Mode III fracture. \textbf{b}, Bamboo fracture modeling with stress field propagation between fibers. Points $P_1$, $P_2$, and $P_3$ represent left fracture height on consecutive fibers $f_{i-1}$, $f_i$ and $f_{i+1}$, with angles $\theta_{i-1}$ and $\theta_i$ showing fracture directions of fibers $f_{i-1}$and $f_i$. Blue dotted circles illustrate the stress field emanating from point $P_2$, determining the probability distribution of potential fracture paths.}
\label{fig:physical_model}
\end{figure}

\clearpage
\renewcommand{\thefigure}{E3}
\begin{figure}[H]
\centering
\includegraphics[width=1\linewidth]{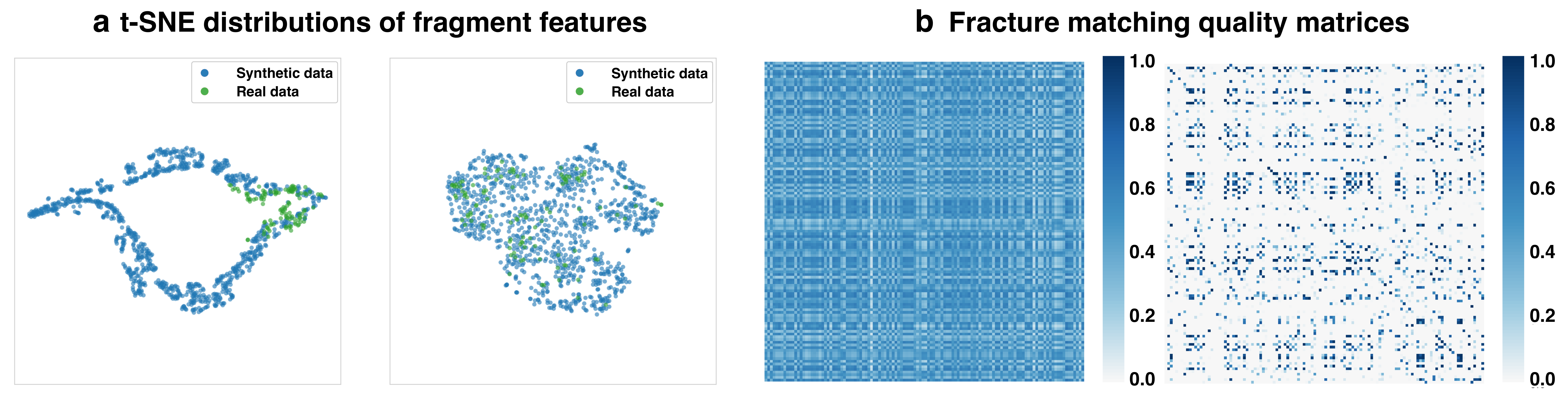}
\caption{\textbf{Data distribution analysis and matching performance.} \textbf{a}, t-SNE distributions of fragment curve features showing the relationship between synthetic data (blue) and real bamboo fragments (green) across two parameter configurations. Convergence indicates effective parameter optimization in model. \textbf{b}, Fracture matching quality matrices for 118 real fragment pairs. Deeper blue indicates higher similarity, with diagonal elements representing correct matches. The lower-right matrix shows good discriminative performance with clear contrast between prominent diagonal (blue) and off-diagonal elements (white), corresponding to the most effective parameter setting (right of \textbf{a}).}
\label{fig:extended_data_parameter_optimization}
\end{figure}

\clearpage
\renewcommand{\thefigure}{E4}
\begin{figure}[H]
\centering
\includegraphics[width=1\linewidth]{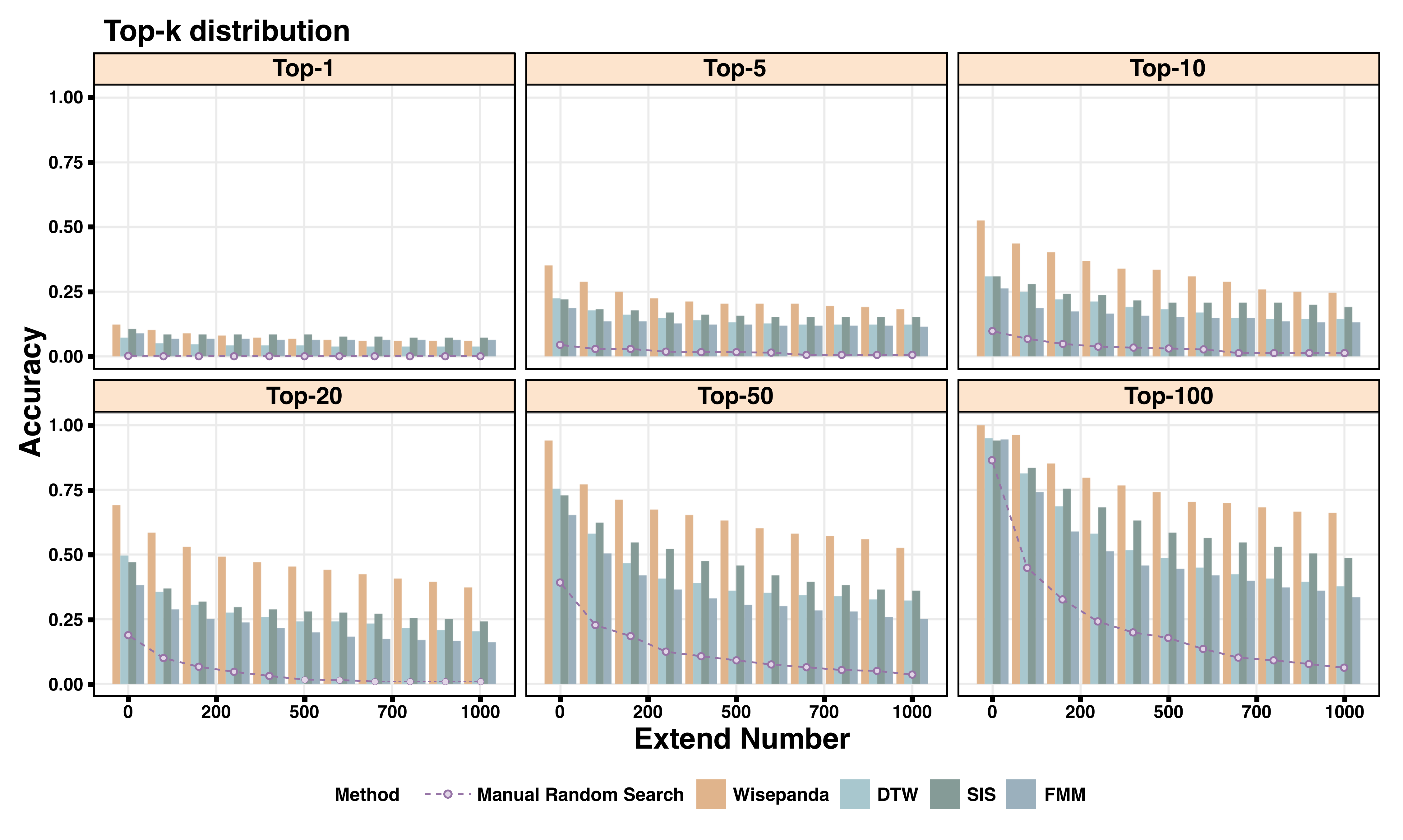}
\caption{\textbf{Robustness analysis of different fragment matching algorithms with increasing interference fragments.} Performance comparison across six accuracy levels (Top-1 to Top-100) as candidate pool expands from 0 to 1000 fragments. WisePanda (orange bars) maintains relatively stable performance despite increasing interference, while Manual Random Search (dotted purple line) shows rapid accuracy decline. Other methods (DTW, SIS and FMM) show lower performance compared to WisePanda across different interference levels. At maximum interference, WisePanda maintains 50-65\% accuracy at Top-50 and Top-100 levels, while competing methods achieve below 36\% at Top-50, demonstrating the physics-driven approach's advantage in real archaeological scenarios with thousands of potential matches.}
\label{fig:extended_robustness}
\end{figure}

\clearpage
\renewcommand{\thefigure}{E5}
\begin{figure}[H]
\centering
\includegraphics[width=1\linewidth]{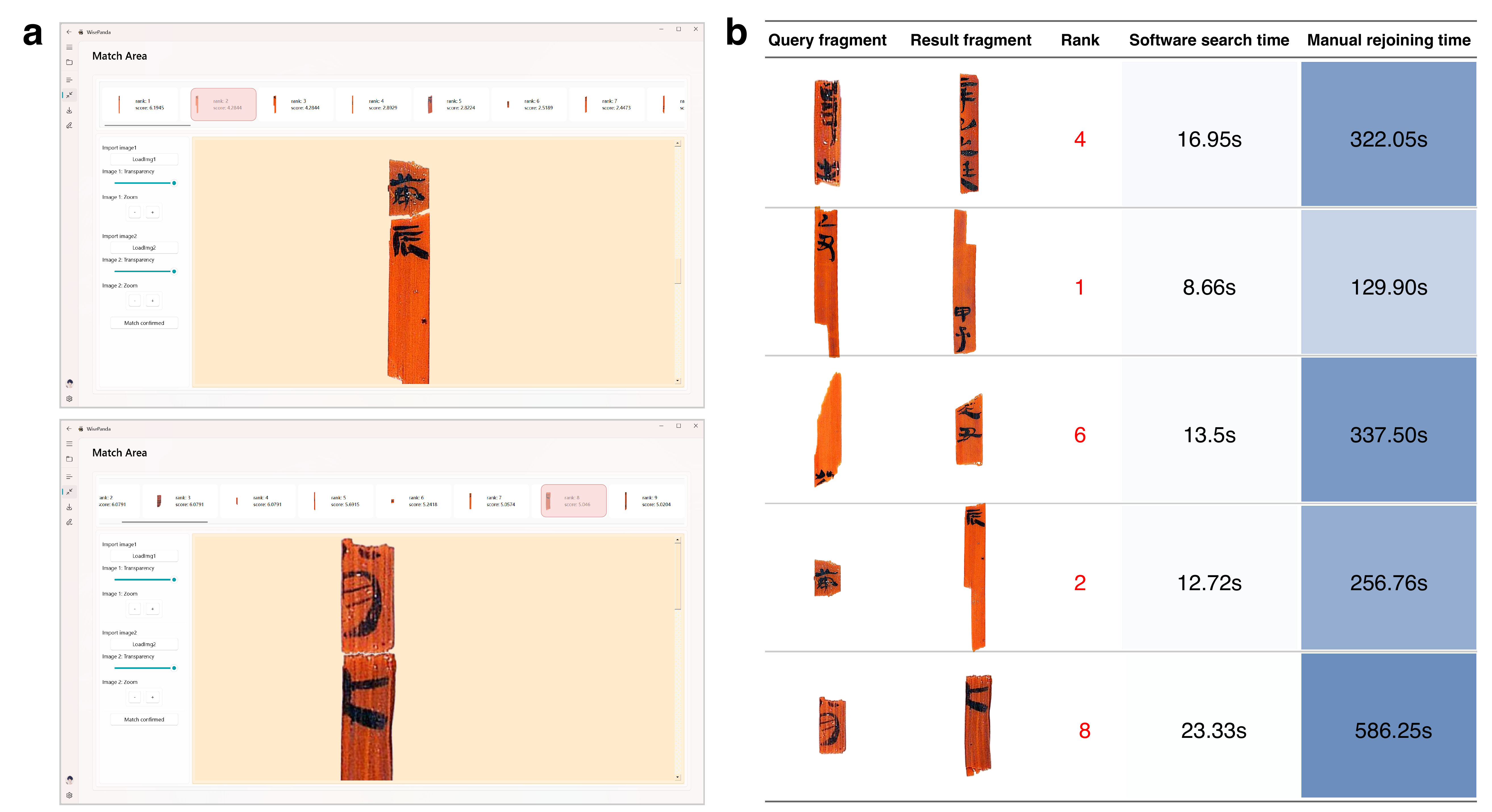}
\caption{\textbf{Practical application of WisePanda.} \textbf{a}, WisePanda software interface showing two successful matching cases: target bamboo slip fragments with their corresponding matching results, demonstrating the system's verification capabilities. \textbf{b}, Efficiency comparison between WisePanda-assisted and traditional manual matching across five test cases. The table records query (top fragment) and result fragments, system recommendation rank, and time measurements for both methods. WisePanda reduces average matching time from 326.49 to 15.03 seconds—approximately a 20-fold efficiency improvement. These results validate the physics-driven approach's practical value for large-scale bamboo slip fragment rejoining.}
\label{fig:software}
\end{figure}


\end{document}